\documentclass{article}
 \usepackage[final]{neurips_2018}
\usepackage[utf8]{inputenc}
\usepackage{graphicx}
\usepackage{caption}
\usepackage{natbib}
\setcitestyle{numbers}
\usepackage{hyperref}
\graphicspath{ {./} }
\usepackage{titlesec}

\begin{titlepage}
\title{Real Time Multi-Object Detection for Helmet Safety}
\author{%
  Mrinal Mathur\thanks{Department of Computer Science, Georgia State University, Atlanta}\thanks{$\{mmathur4,abenkkallpallichand1,vnuthalapati1\}@$student.gsu.edu},\\
  Archana Benkkallpalli Chandrashekhar\footnotemark[1]\footnotemark[2], \\Venkata Krishna Chaithanya Nuthalapati\footnotemark[1]\footnotemark[2]
 }

\end{titlepage}
\begin{document}
\maketitle 

\begin{abstract}
    The National Football League (NFL) and Amazon Web Services \cite{NFL} (AWS) teamed up to develop the best sports injury surveillance and mitigation program via the Kaggle competition. Through which the NFL wants to assign specific players to each helmet, which would help accurately identify each player's “exposures” throughout a football play. We are trying to implement a computer vision based ML algorithms capable of assigning detected helmet impacts to correct players via tracking information. Our paper will explain the approach to automatically track player helmets and their collisions. This will also allow them to review previous plays and explore the trends in exposure over time.
\end{abstract}

\section{Introduction}
The National Football League is America's most popular sports league. Founded in 1920, the NFL developed the model for the successful modern sports league and is committed to advancing progress in the diagnosis, prevention, and treatment of sports-related injuries. Health and safety efforts include support for independent medical research and engineering advancements as well as a commitment to work to better protect players and make the game safer, including enhancements to medical protocols and improvements to how our game is taught and played. This work was done for analysing sports injury and mitigating this problem by surveying them. there are few novel works before where they detected the helmet impacts, but here they are taking it to the next level by assigning each player to each helmet and accurately identity player's exposure throughout football plays

Currently, NFL annotates each subset of players each year to check the exposure and to expand this feature they require the field view to determine each and every players positions. This competition will be a subset of the previous year's competition from kaggle.com, and we will be using almost same dataset with more information given to use on dataset. Along with information, each player is associated with videos, showing a sideline and endzone views which are aligned so that frames corresponds between the two videos. To aid with helmet detection, we are also provided an ancillary dataset of images showing helmets with labeled bounding boxes. This year they also providing baseline helmet detection boxes for the training and test set. As per the rule after submission of this code, they will  run the submitted model in 15 unseen plays that is a part of a holdout training set. During the competition participants are evaluated on the test set and a part of the predictions is used to calculate the above mentioned metrics and show the best score for each participant on a public leaderboard. After the competition the score with respect to the remaining part is released and used to determine and display the final scoring (private leaderboard). For training our models we used Pytorch with object detection algorithms with tracking. Moreover, we use several implementations and pretrained weights. Our training code is available online \cite{NFL_github} Our paper will discuss on what methods we implemented to detect these players with helmet collision and how accurate it was. We will explain the paper in following manner: Exploratory Data analysis, Methodology, Inference and Conclusion

\section{Exploratory Data Analysis}
The main task before we even start our modelling part is to first understand our data and infer what we can extract from that information. First we looked into the basic information on what data is being given to us and how we will represent those data. we came across information like the bounding box information has been provided for each player. Our main training data consist of player\_id, x-y coordinate of those players, direction of the players,the snap information, team information and which frame they belong to. 
While trying to understand the training and testing information, we found that testing data is a subset of training information. We have 120 training videos out of which 60 are sidezone and 60 are endzone images \cite{NFL}. Does Sideline and Endzone Video has same frame? No, 25 plays out of 60 doesn't match and the difference is mostly 1 frame but there are 7 frame difference also. We checked even the biggest and smallest size of the helmet, that means that with respect to the frame, how much percentage is being taken by an individual helmets. 

\begin{figure}[h]
\centering
\includegraphics[width=75mm]{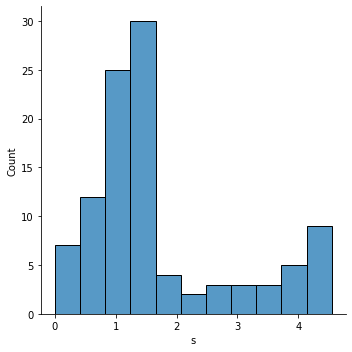}
\caption{Counts of snap frame where movement is the highest}
\label{fig:snap1}
\end{figure}

\begin{figure}[h]
\centering
\includegraphics[width=100mm]{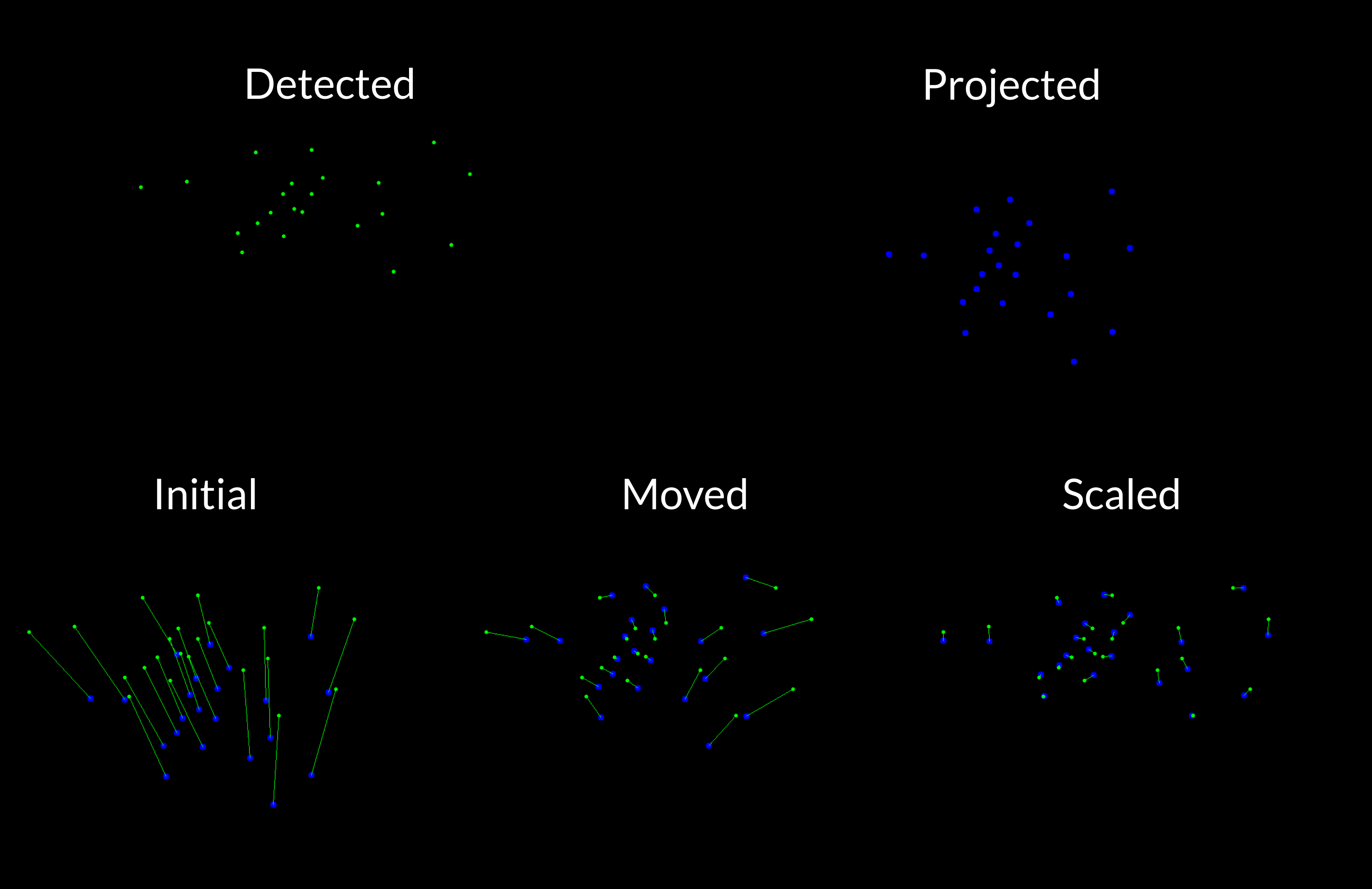}
\caption{Cloud Point Information with detected and projected view}
\label{fig:cloud_point}
\end{figure}

Since each video started with players starting before the match we need to consider which frames we need to consider so that we can extract maximum features, we realised that the game only started after the "snap", Before snap players change their position a bit but after snap they start to run, so we cut the frames before the snap. As seen from the Figure \ref{fig:snap1}, that the movement after snap is highest. 
\\
We can find all the information on our data Exploratory in Appendix A. Since we are talking about players and their movements, we need to check how to track those players. We used tracking information by extracting each images using the ffmeg tool that extracts the image and put it into the memory and after loading it the photo will be deleted from the memory, which will reduce the overhead of uploading more than 90,000 images into the memory. After extracting these images, we use the data from "tracking\_baseline\_player.csv", where we were given x, y coordinate of players, their orientation "o", and their speed and acceleration. As shown in Appendix A, we were able to extract information on orientation and their angles which gave us an idea on how to track the camera mapping as well, we located the global camera position $(x, y, z)$ and adjusted their rotation $($yaw, pitch and raw$)$ by using Iterative Closest Point Algorithm [reference]. Since we needed the initial rotation angles, we used euclidean distances between the points from tracking of 2D projections and detection point clouds after transformation, this was useful since the points to be matched are chosen from all four direction points and consider it as the main cluster as shown in Figure \ref{fig:cloud_point}

This helps in measuring the approximate distance between the camera and the actual position of players.
\section{Implementation}
\begin{figure}[h]
\centering
\includegraphics[width=100mm]{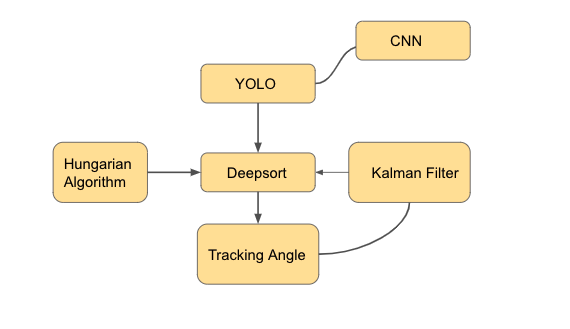}
\caption{Top-level Model Pipelines}
\label{fig:pipeline}
\end{figure}

Since we are working on the multi object detection, we started using a model that works really powerfully while detecting the images, hence, we used YOlOv5 for our case, but along with this we also wanted to use  algorithms for tracking the players in our case. This gave rise to a combination of YOLOv5 and DeepSort algorithms. In this section we will be discussing about various methods that we used to make our final model work.
our solution consist of the following stages as shown in Figure \ref{fig:pipeline}: 
\begin{enumerate}
    \item Use YOLOv5 for detecting helmets.
    \item Use deepsort to track the helmet.
    \item  Tracking Players: 
    \begin{enumerate}
        \item Use ICP (Iterative Closest Point) and the Hungarian algorithm to assign helmet boxes to the tracking data for each frame.  
        \item Divide helmets into two clusters using k-means based on the color of the helmet.
        \item Use ICP (Iterative Closest Point) and the Hungarian algorithm to assign helmet boxes to the tracking data again, taking into account the results of deepsort and the helmet color information.  
    \end{enumerate}
    \item Use the Hungarian algorithm to determine the final label. 
\end{enumerate}

\subsection{Stage-1: Detecting Helmets: YOLOv5}
We used only those frames for YOLOv5 that are used after the snap of the plays, we trained YOLOv5x6 with an image size of 1280. After converting the videos to images, we trained more frames with impact.
We used both the sideline images and endzone images but, we flipped the endzone images and used it for augmentation of the images for better extraction. Since helmets with impact are often only partially visible, we had to set lower thresholds for confidence and IOU \cite{zhu2020unpaired} along with DeepSort 
\subsection{Stage-2: Use deepsort to track the helmet} we used DeepSort algorithm as two main extracting features for each helmet box from Deepsort feature extractor using the pretrained weight (ckpt) and save it to disk for using Hungarian algorithm later. Deepsort mainly consists of 2 algorithms. They are Kalman  filter and Hungarian Algorithm. Kalman Filter helps us to track the helmets and also helps us to estimate the exact position of helmet even when the detected Helmet in frame Number N  gets hidden in frame number N+1.  We used this mostly to reduce false positives, we deleted helmets that could not be tracked for frames longer than a certain frame, for example, if a particular person is not moving in 5 frames, we will store the information in memory for that player and update it only when they move, this way we modified the deepsort  to return the features extracted by id and the confidence. These extracted features are used in Hungarian Algorithms and ICP, and confidence is used in tracking the players and mapping them into the X-Y latent space.
\subsection{Stage-3:  Tracking Player}
We used three  methods to make sure we track the players using their angle and rotation information and labeling each of them into a latent space:

\subsubsection{Iterative Closest Point \& the Hungarian algorithm } 
\begin{figure}[h]
\centering
\includegraphics[width=100mm]{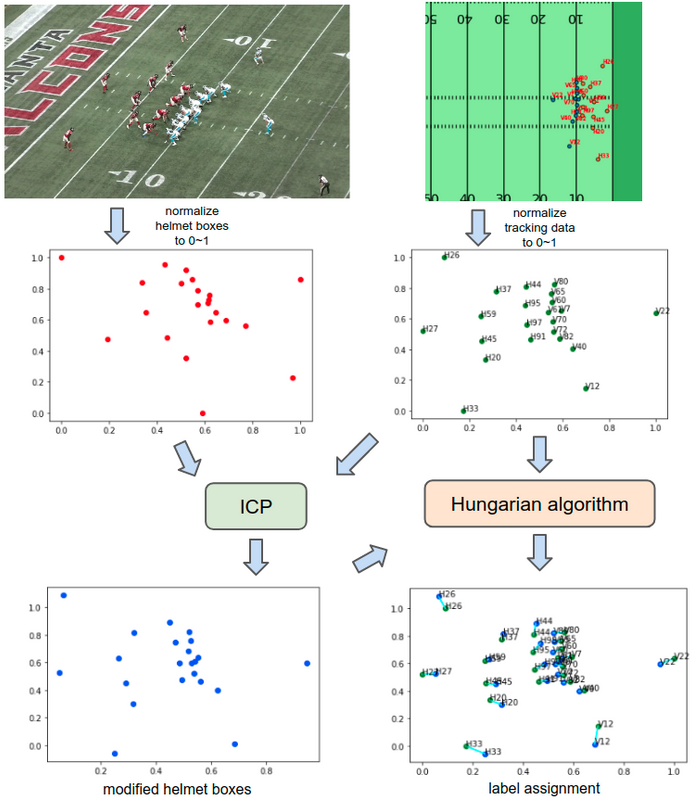}
\caption{ICP and Hungarian Mapping}
\label{fig:icp_hun}
\end{figure}

As per  \cite{10.1117/12.57955} Iterative Closest Point (ICP) is used finding the transformation between a point cloud and some reference surface, in our case that the player's information. We used ICP to adjust the position of the helmets bounding boxes so that the total distance between the helmet and tracking data is minimum, Before we even applied our ICP, we normalized the players into X-Y coordinates and then we apply Hungarian algorithm on each tracking data. We first calculated the shape context for each points in tracking data and calculated the cosine distance matrix between this sensor information and bounding box information as shown in Figure \ref{fig:icp_hun}. Then Hungarian algorithm uses the preceding information to assign each helmet's boxes for tracking data. Theses labels are assigned by creating again by cosine similarity cost.

\subsubsection{Divide helmets into two clusters} 
\begin{figure}[h]
\centering
\includegraphics[width=100mm]{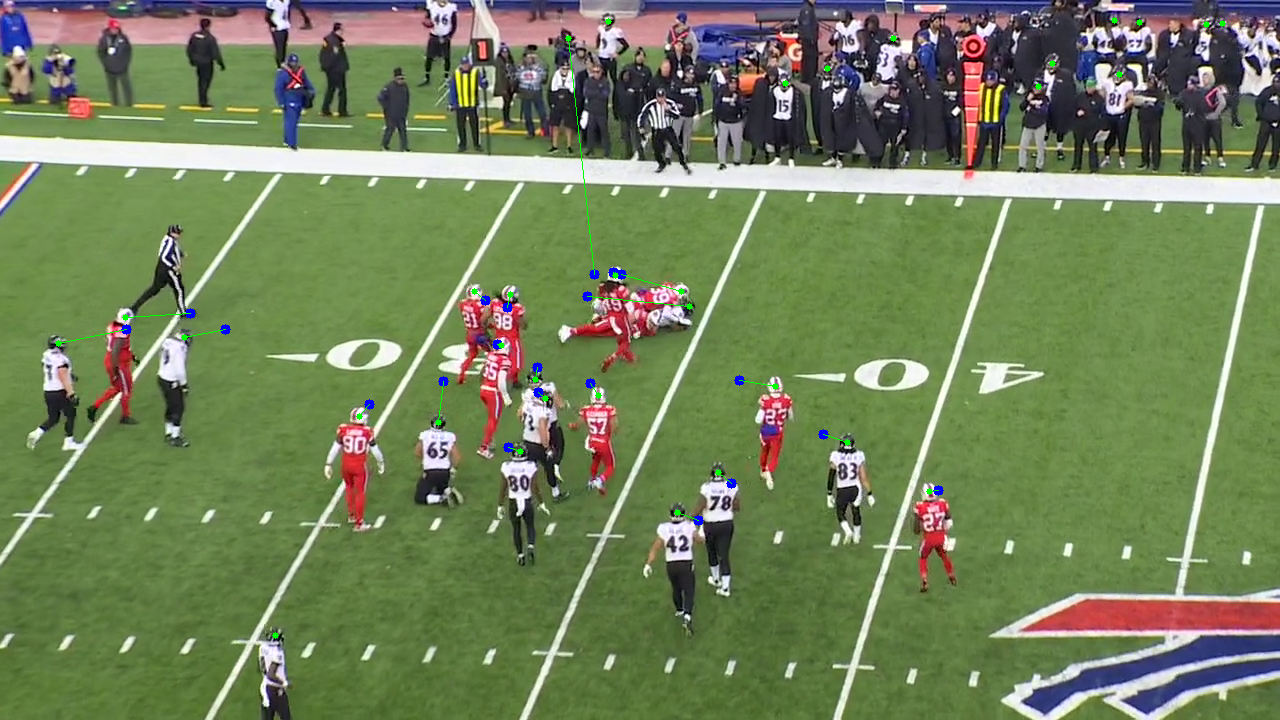}
\caption{KMeans Figure}
\label{fig:kmeans}
\end{figure}

as seen in Figure \ref{fig:kmeans}, We divide these helmet boxes into two clusters based on the color of helmets using kmeans, and calculated the distance between each helmet and used the features calculated by deepsort, this we can combine with above process and determine which team the helmets belong to.

Sometimes, it is necessary to disable detections that doesn't belong to any tracking data but are connected to some missing points anyway. First of all, we try to disable a single outlier if it is. If it fails, we try to find a main cluster by border points as it was described earlier (up-down-left-bottom points). It helps us to disable all the points that don't belong to the main cluster. 
\subsubsection{Use Homography, ICP and Hungarian algorithms} 
\begin{figure}[h]
\centering
\includegraphics[width=150mm]{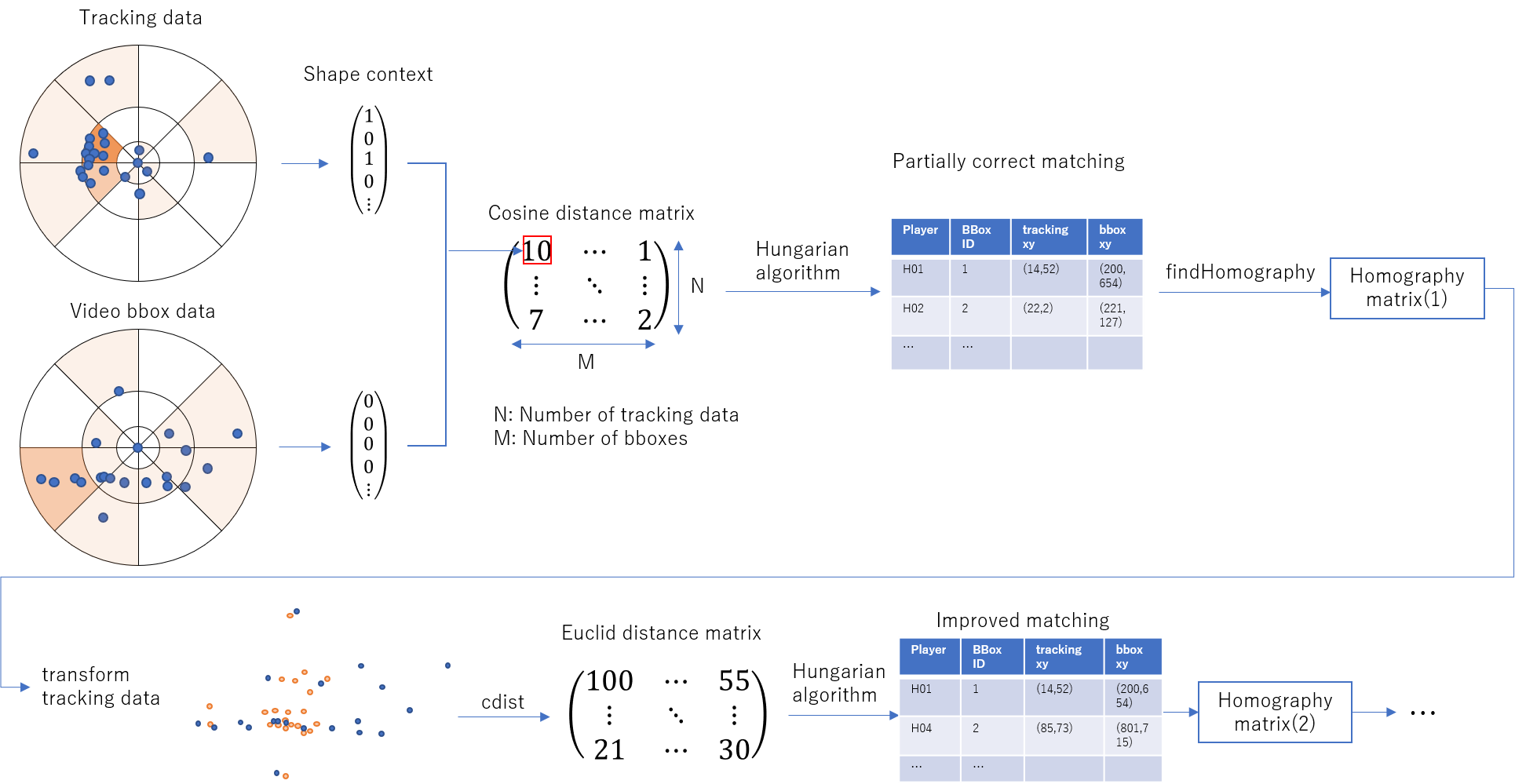}
\caption{Tracking using Homography and Hungarian Algorithm}
\label{fig:tracking}
\end{figure}

We modify homography transform in a preceding frame to get that of current frame Since player positions of adjacent frames are similar, the homography matrix should be also similar. First, we applied homography transform in previous frame to the tracking data in current frame as shown in Figure \ref{fig:tracking}. Then we assign each players to bboxes(conf$\geq$0.40) in current frame like aforementioned way to get current homography matrix. To improve robustness of the algorithm, we tried homography matrices in 60 preceding frames and selected one with best cost. To deal with false positive of helmet detections, we had to add rows full of zero in distance matrix (ignore false positive bboxes). The number of ignored helmets was adaptively selected by looking at cost function. If increasing ignore number from N to N+1 helped decreasing cost function by 10, the increase was accepted.
\\
\subsection{Final Label}
Given in the above Figure [figure name], we created the cost matrix based on percentage of labels in the same deepsort helmets. Therfore, the labels that appear frequently in the same deepsort helmets are assigned priority and the image features are reused from K-means team clustering.

Since there are often less than 22 players in the video, we adjusted the number of players by removing the players on the top, bottom, left and right of the tracking information (by cutting off the tracking coordinates with a rectangle), and then solved it as a linear assignment problem. There are many candidates for removing players, but we chose the candidate with the smallest minimum distance among all candidates.
\subsection{Training}
A pipeline of 3 models are designed to reach our goal. YOLOv5, DeepSort, Angle Tracking. Each is trained on the data that helps the model learn to predict what is expected out of it.

Images are extracted from the video each frame being an image, these are collaborated with the other data available in CSV, such as position, bounding box and label to train each model in the pipeline.

As a machine learning model gets efficient with more data available, and everyone has access to exact dataset, to get an edge over other contenders we came up with an idea.  The videos are sideLine or endZone which are two different points to capture a video.  Models need to learn better at tracking the players. We have flipped the EndZone x,y coordinates along with the tracking data. This gave us an extra set of data for models to learn from the data. The flipping has kept the players movement and game play intact. The same has not been done on the sideline because a lot of frames in sideline have outsiders and audience, This throws the model from focusing on the players and takes away the focus from the players.

\begin{itemize}
 \item \textbf{Bounding Box(YOLOv5):}  Images are extracted from the videos taking each frame as an image. The player's bounding box and label is available in the CSV. This data is collaborated into an annotation file of YOLOv5. The YOLOv5 is trained to learn to put labels and bounding boxes on the new images given to it, similar to the images that it is trained on.
The clusters of helmets(bounding boxes) are created across the frames in the video by applying the trained model on each frame in the video, the most common label in the set of frames will be selected and labeled with that name.

\item \textbf{Tracking(DeepSort):}  Image being labeled and setting the boundary is coupled with tracking it across the frames, making us work on a video rather than images. The model is trained in a sorted frame from 0 to N, which is the start of the game to the last frame. The images are coupled with the players location and coordinated synced with the video frames. The series of images help the model to learn to predict the exact location of the player in the next frame. The DeepSort model’s default parameters that will impact the performance, and helps us track our players in our problem are selected and altered to help us track the players efficiently. Many frames have players overlapped, hidden or cropped off at the edges making their location not clear in the image extracted from the video, Kalman filtering helps us handle the occlusion between the frames that can happen in the game play. This improves the tracking and fills the gap in the frame while tracking the  players throughout the gameplay.
The Kalman filter working can be referred here \cite{KalmanFilter} 

Comparison of the frames and to understand the players positions and to Hungarian algorithm is used to solve the Assignment problem. It is an efficient way of comparing the data available in matrix form. A quick reference on how this helps in tracking the players can be read here: \cite{Hungarian_Algorith}

The usage of the YOLOv5, and DeepSort to detect and track objects in a video is based is similar to the research paper \cite{DBLP:journals/corr/WojkeBP17}

\item \textbf{Collision Detection(Angle Tracking):} Previous models help us label and their bounding boxes across frames. Using the speed, direction, and distance of two players can help us predict the collision. Players previous, current, next location in the frames is available to us from the csv data. It  is used to calculate the angle, and use it as a metric to detect the collision. The calculation of the angle is done with homography of the image with the next and previous tracking info of the players. Gaining the motion of the player and his direction. Using the angle of their motion and location we can trigger the event of collisions. The model is trained with images in sequence out of the video with the tracking information.
Homography is a matrix transformation that is applied on the image with respect to a point to orient the image focusing on the point. Applying this on an image based on tracking positions of a player from previous, current and next frame will help us track the player in gameplay. \end{itemize}

\subsection{Testing}
Each model is tested as an individual model once each model is performing as expected they are tested at once for the task on hand. The output of each model is cross-validated on the pool of data to check their performance. As we know that the testing data is a subset of training data, the chance for us to overfit the models is very high. To prevent this  cross validation is used to test the models and their collaborative output. 

YOLOv5 output throughout the frames in a video is segregated and any duplicate labels or frames data is dropped and formatted into the csv format for the submission.

Once the models are in the place we have explored the parameters for the model to find the optimal combination for the best accuracy. A pattern of gradual increase and decrease in the accuracy was observed in the combinations. 

Kaggle Submission: The kaggle has a set of instructions on the output for our model.
 The helmets that are involved in an impact are 1000x more weighted to calculate the accuracy metric when we are testing evaluating the model. 
We have testing on the metric of having a limit of 22 helmets per frame, and ignore the players on the sideline(non players).
The positions predicted by the model shall be in the limits of the video frame dimensions. 

The submission will be executed on 15 unseen gameplay videos that are not available to us, and all the training set given to us is a subset of the training data. This needs a careful method of handling the model training to prevent any overfitting of the data for which we have used cross validation.

\subsection{Evaluation}
In the kaggle competition, One of the metrics considered to evaluate our Submission is the IOU and weighted accuracy which are defined as:\\
$$IOU = \frac{A\cap B}{A\cup B}\geq0.35$$ 
$$Weighted Accuracy =\frac{TotalCorrect_{nonimp}+TotalCorrect_{imp}*1000}{TotalHelmets_{nonimp}+TotalHelmets_{imp}*1000}$$
where: \\
$TotalCorrect_{nonimp}$ is the number of correctly assigned non-definitive impact helmet boxes.\\
$TotalCorrect_{imp}$ is the number of correctly assigned definitive impact helmet boxes.\\
$TotalHelmets_{nonimp}$ is the total number of non-definitive helmets boxes.\\
$TotalHelmets_{imp}$ is the total number of definitive helmet impact boxes.\\

Our maximum Leaderboard score achieved so far from the competition is 0.713(Private) and 0.797(Public).\\

\begin{figure}[h]
\centering
\includegraphics[width=150mm]{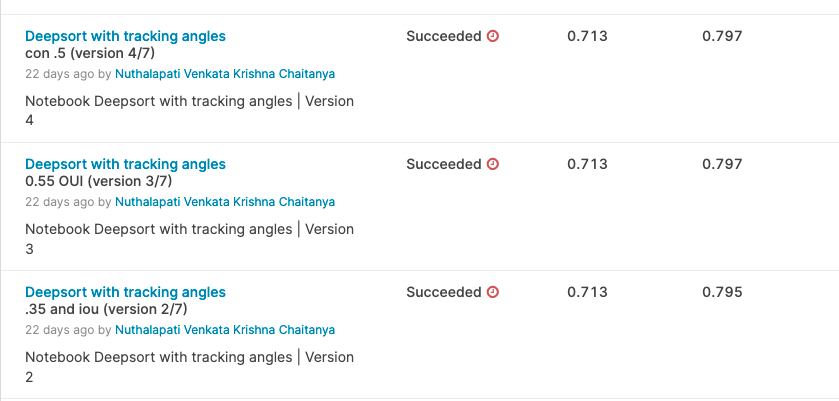}
\caption{Kaggle Submissions}
\label{fig:kaggle}
\end{figure}

We performed evaluation on object detection capability of YOLOv5 model. As a part of future work, we could work on evaluation of collision detection.
\href{https://wandb.ai/archanabc/YOLOv5/runs/3hfrmrzo?workspace=user-}{Click here to view Wandb workspace} 
\subsubsection{Setup}
We are using Weights and biases tool \cite{W&B} to visualize the experimental metrics of our model, as it is configured with YOLOv5 and provide us with metric tracking of the model performance while training.
We have split the extra images (data/images) into training , validation sets. As we need to test the model with new data to correctly determine best parameters for data using validation data set.

\subsubsection{Results}
\begin{figure}[h]
\centering
\includegraphics[width=\textwidth]{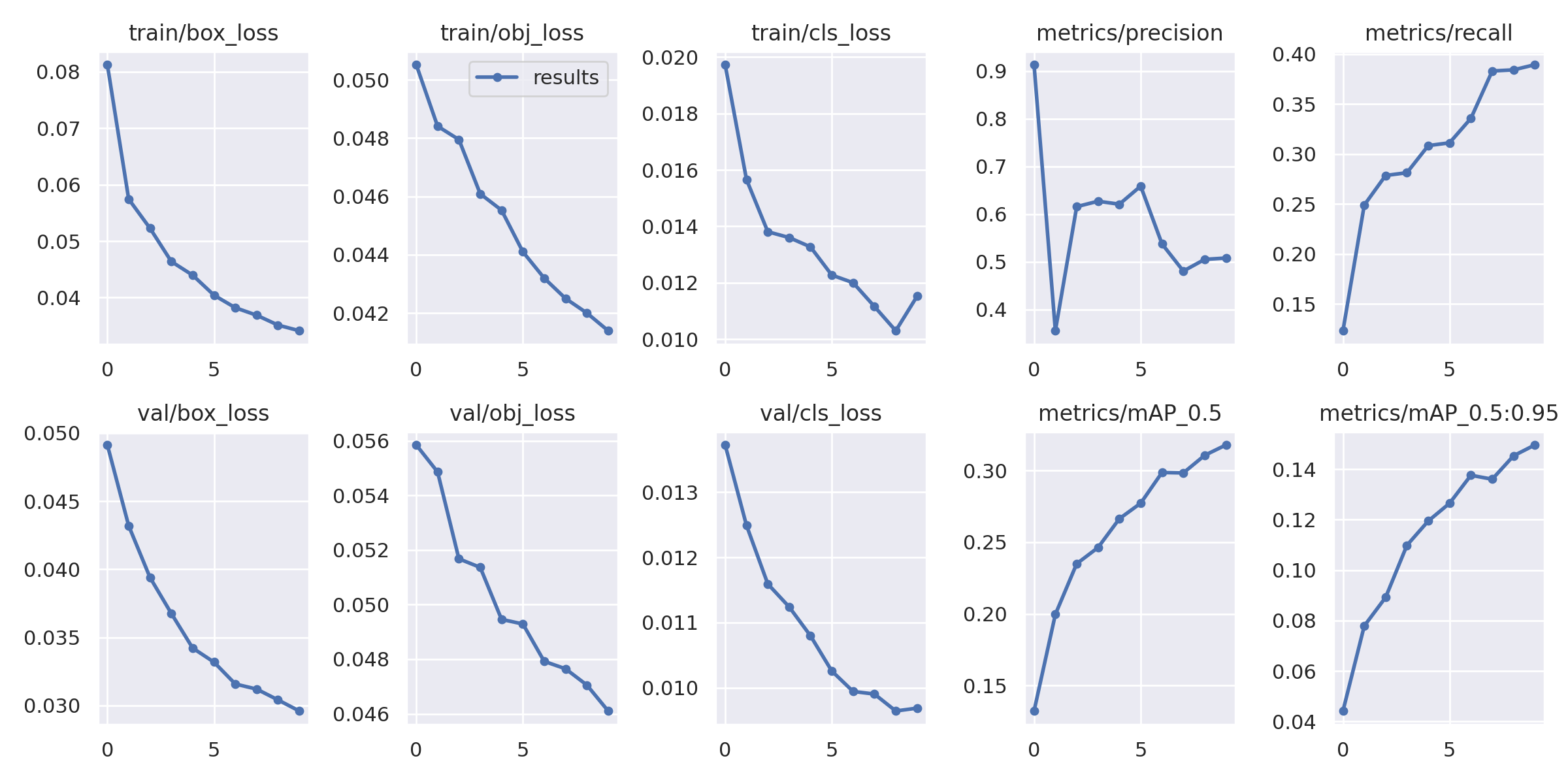}
\caption{Overview of losses over train and validation data set for 10 epochs/cycles.}
\label{fig:losses}
\end{figure}

Below Figure \ref{fig:losses} shows prediction loss metric over the training and validation set losses , precision, recall values over the training and validation set and mAP(mean Average Precision) to analyse object detection . 
box\_loss is the regression loss is the inability to detect helmet. Obj\_loss is the object loss that determines objectiveness of prediction with respect to ground truth. cls\_loss is the classification loss due to misclassification. This gives better understanding of the performance of the model. As the model gets trained with new data in each cycle of training, the prediction of bounding box losses reduced. The generated report is available at \cite{W&B_Results_Visual} and Kaggle run output at \cite{W&B_Git}.

\begin{figure}[h!]
\centering
\includegraphics[width=75mm]{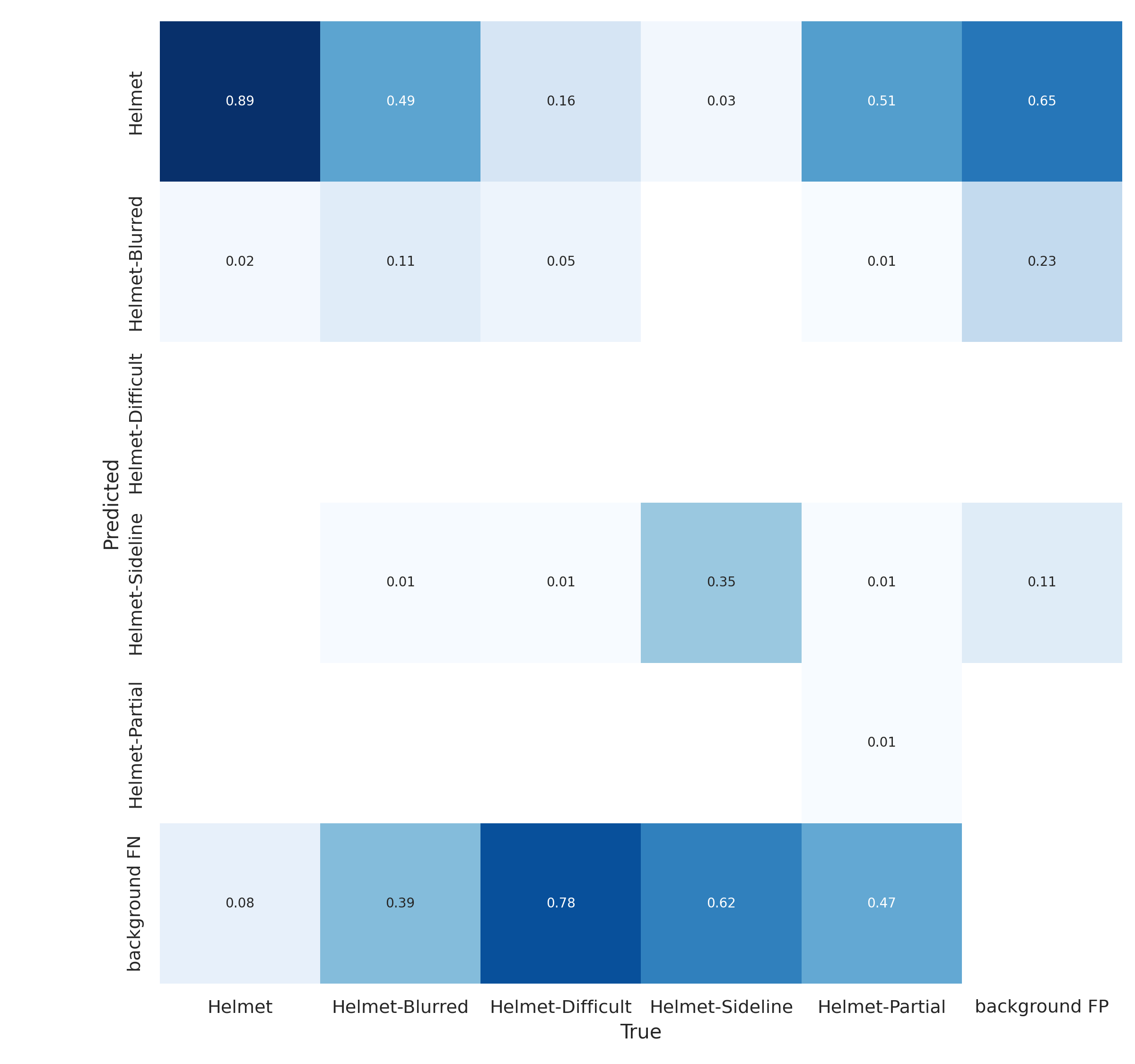}
\caption{Confusion Matrix}
\label{fig:confusion}
\end{figure}

There are 5 possible helmet classes on which our model can apply bounding boxes on. Overview of how our model classifying the objects can be analysed through the summary of confusion matrix for 5 classes as shown in Figure \ref{fig:confusion}. For instance, We can infer that
89\% of the helmet (clearly visible) class are correctly classified out of actual ground truth.\\
 \\

\begin{figure}
\includegraphics[width=70mm]{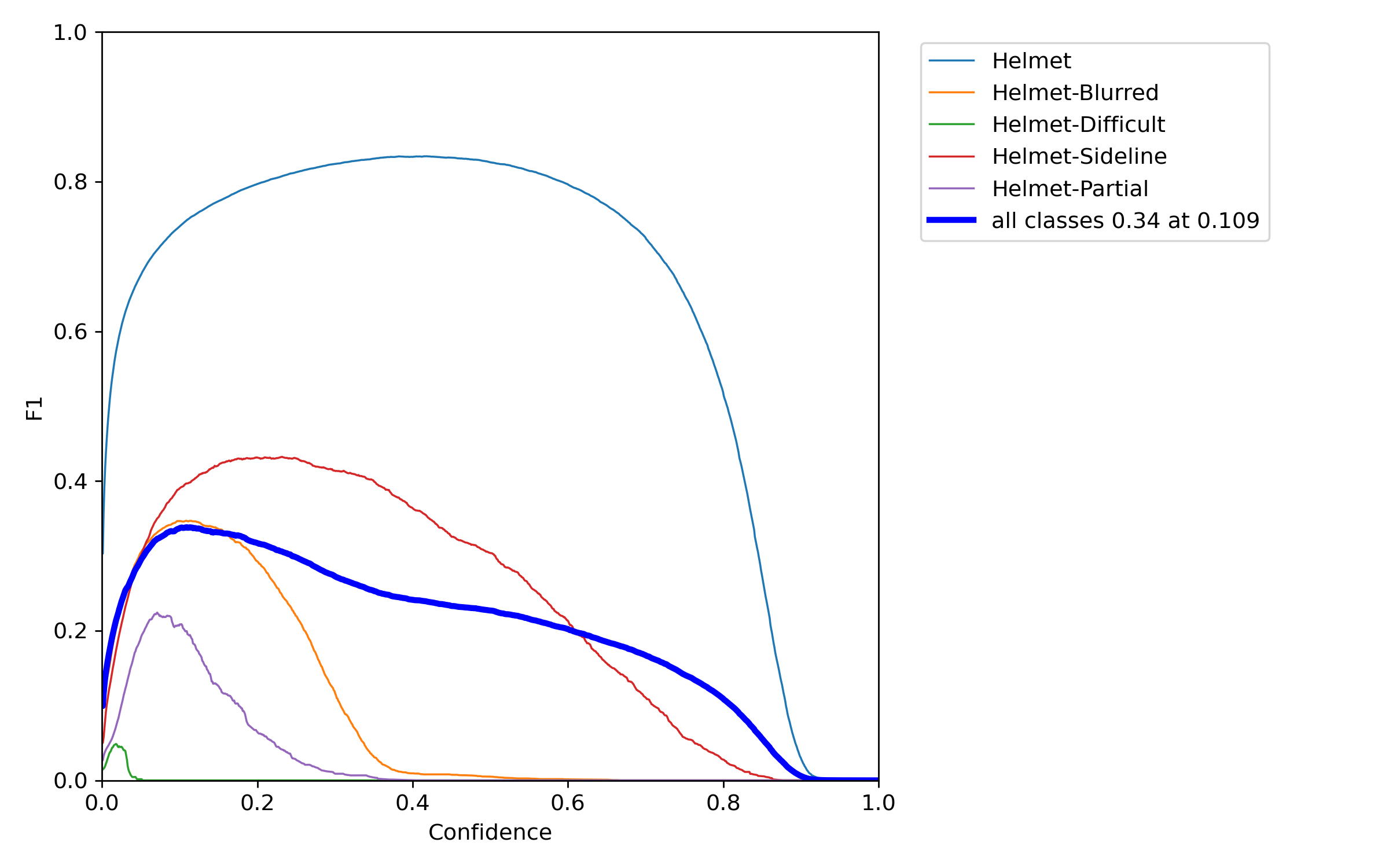}\includegraphics[width=70mm]{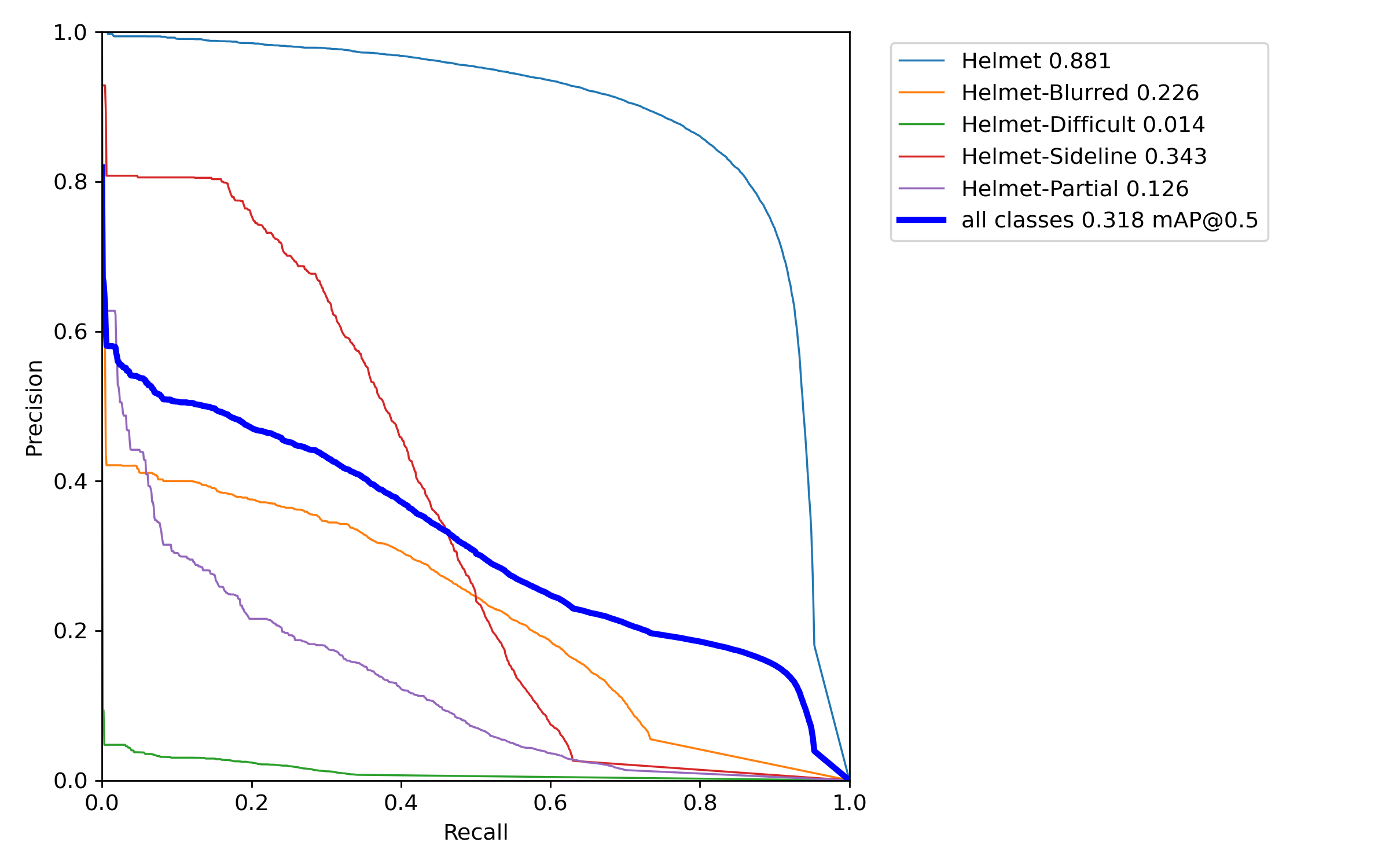}
\\
\includegraphics[width=70mm]{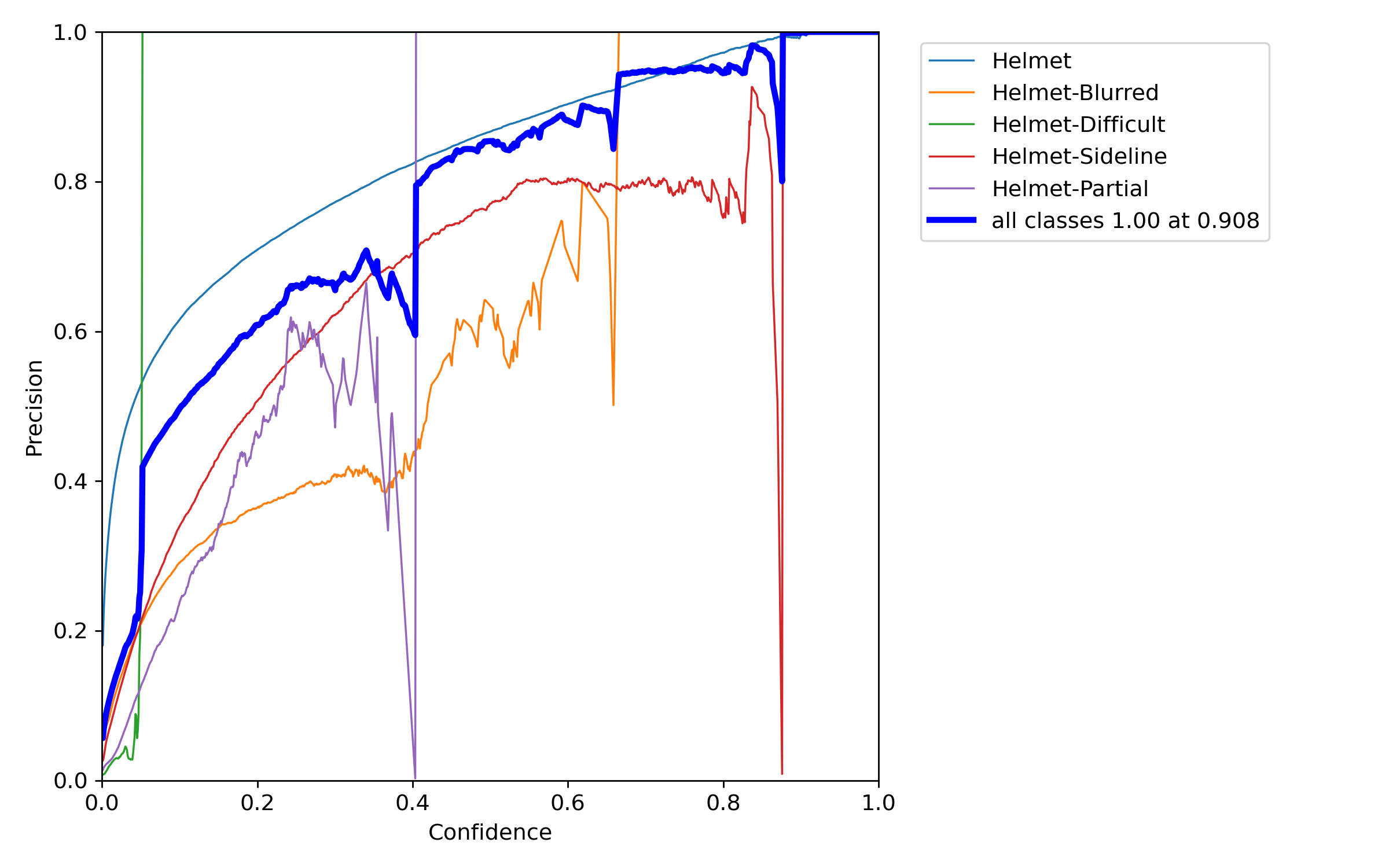}\includegraphics[width=70mm]{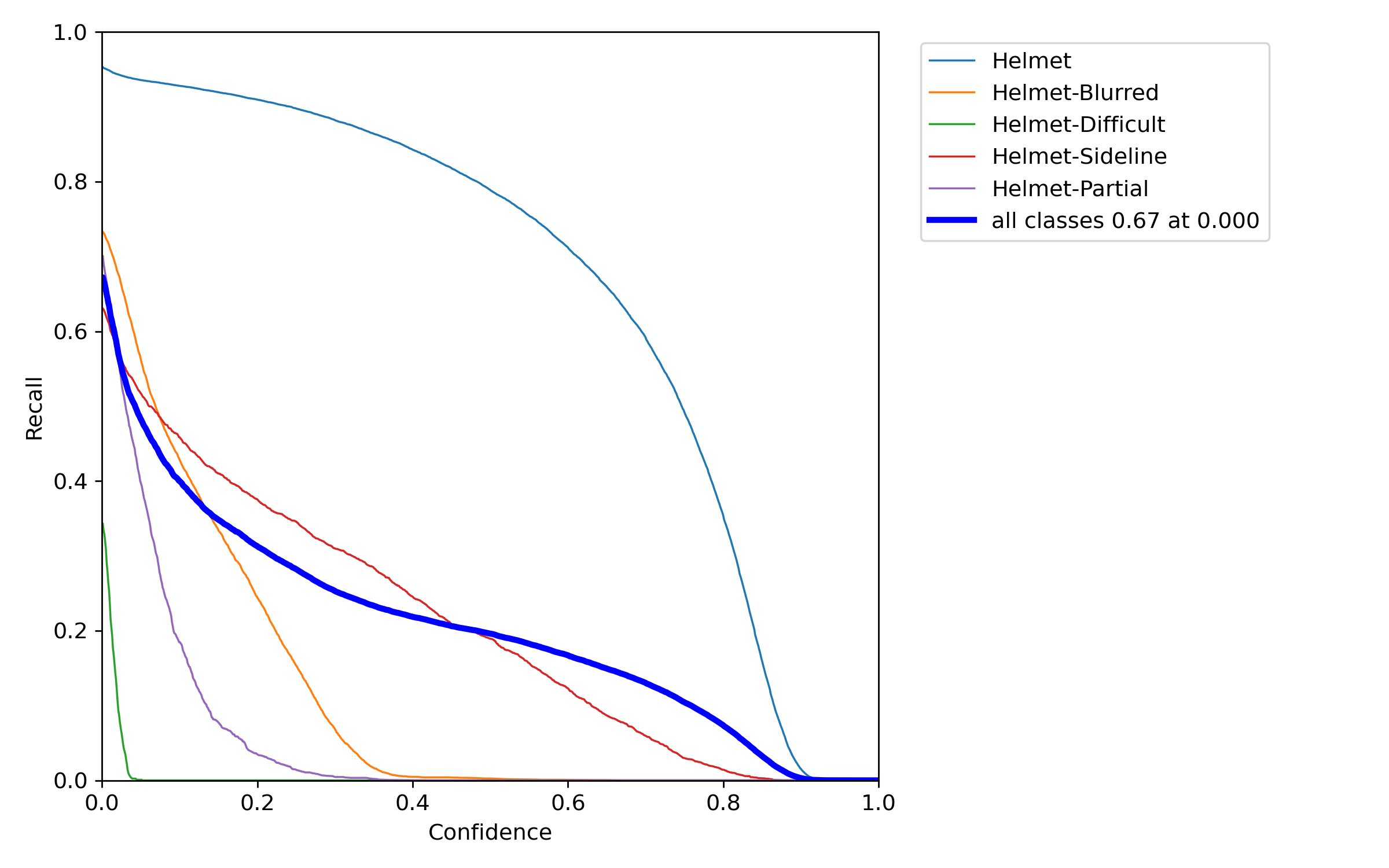}
\caption{Further analysis of Evaluation-helmet visibility based model prediction results.}
\label{fig: l1}
\end{figure}

 We can determine that the helmet detection capability of the model is dependent on these classes (image quality) as shown in Figure \ref{fig: l1}. As per the results shown , we can see that in a given image,to achieve better object detection accuracy, setting the confidence between 0.2 to 0.6 would yield a overall better score for the given data.   
 
\href{https:\/\/wandb.ai\/archanabc\/YOLOv5\/reports\/Shared-panel-21-12-04-13-12-02--VmlldzoxMzAxMzcw}{Click here to view clear image results} 

\section{Conclusions and Future Works}
We introduced you several insights of data by exploring each question and extracted features for our implementation which we described in our ML pipeline. We presented several improvements to our approach for helmet impact detection leading to a position of 34 (private leaderboard). We showed how a combination of tracking the players along with YOLOv5 and DeepSort efficiently leverages over even the slightest movement of players. We confirmed that the application works even better when we used orientation of the players and this works better in case of previous year's NFL dataset as well. We also explained all the elements of our pipeline and explored. We saw that we were able to use an exhaustive pipeline to reach a good enough score. It  would be interesting how we could have used the speed, direction of motion and feed the respective data available for the model to learn better. Also, we would have like to see how Detectron2 would have affected the accuracy of the model

\section{Acknowledgement}
We would like to thank Sumanth and Lauren James for their contribution in project. We sincerely thank our Professor, Dr. Sergey Plis, for his guidance and teachings that led us undertake this project and gain more practical experience on understanding the Machine learning concepts that were taught in class. 

\medskip


\small
\bibliography{sample}




\section{Appendix A}

\subsection{EDA extended}
we will check each an every question that we were able to answer using EDA:

\begin{enumerate}
    \item Get Image by frame:
    \begin{figure}[h]
    \centering
    \includegraphics[width=100mm]{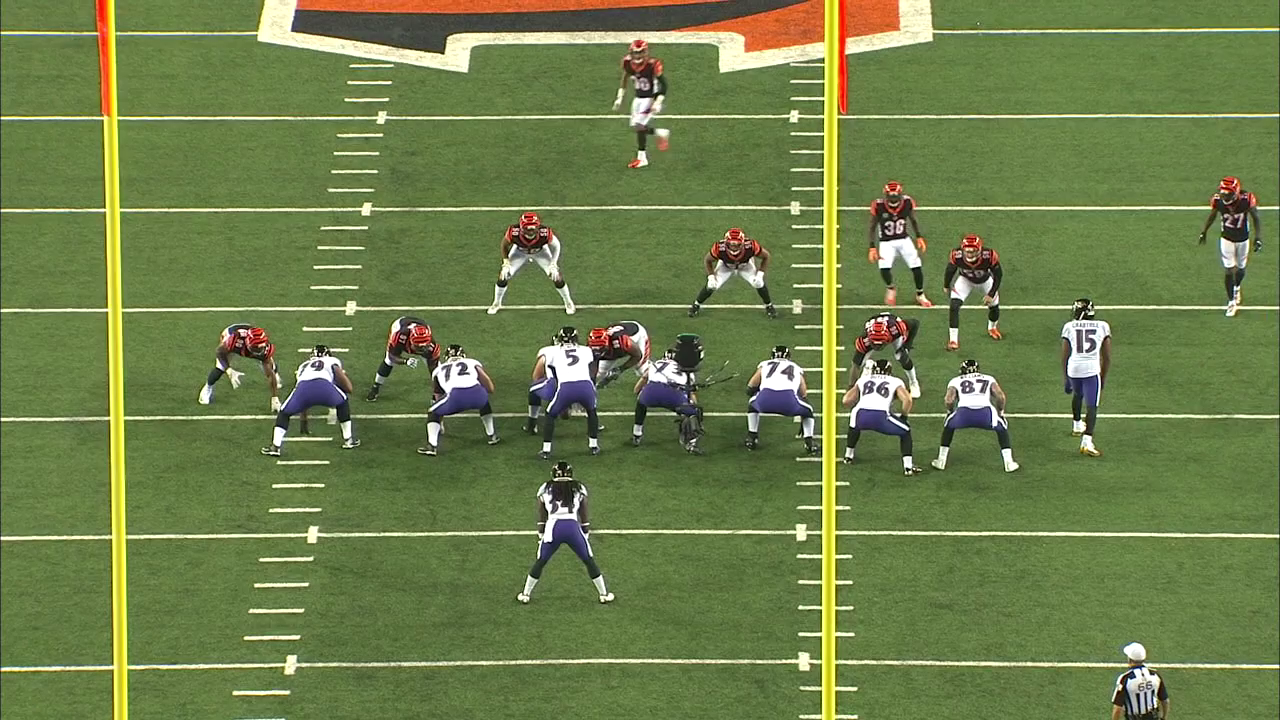}
    \caption{Extracted image}
    \label{fig:detectron2}
\end{figure}
    
    \item How many train videos are there?\\
    Total 120! Sideline and Endzone video pair for every plays!
    \\ \\ 
    train 120 videos
    Endzone 60
    Sideline 60\\
    \item Does Sideline and Endzone Video has same frame?\\
No 25 plays out of 60 doesn't match and the difference is mostly 1 frame but there are 7 frame difference also. \ref{fig:frame}
        \begin{figure}[h]
        \centering
        \includegraphics[width=100mm]{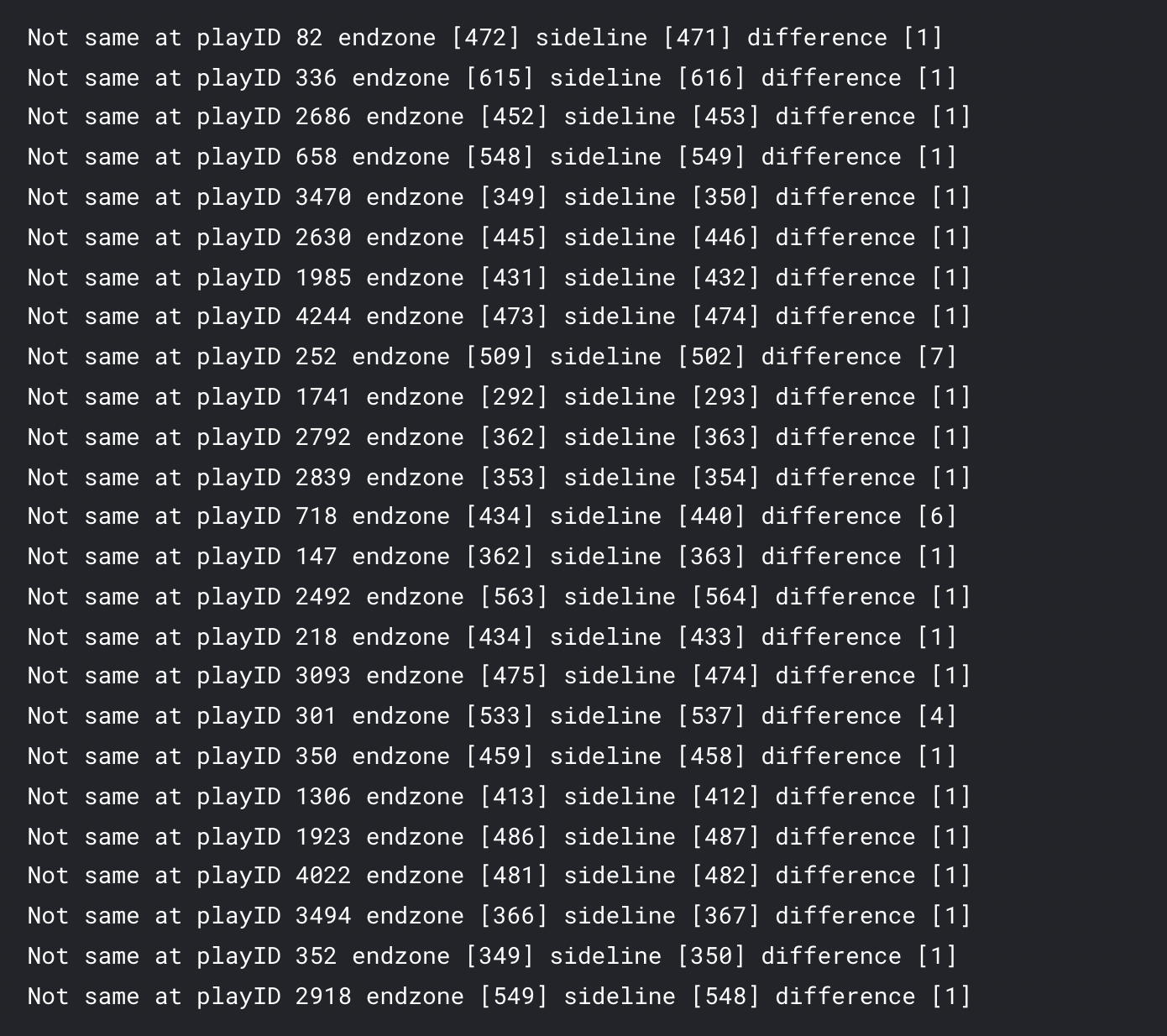}
        \caption{same frame}
        \label{fig:frame}
        \end{figure}\\
    \item Does video frame matches with the frame recorded in the train\_labels.csv?\\
Yes! Frame matches exactly! It's clean!
    \item Is there a frame 0?\\
    
    Yes, there is 1 frame that is 0 and seems mislabeled. So could be simply dropped.
    
    \item How many game, play, frame?\ref{fig:frame_per_game} \\
    gameKey and playID has unique ID\\
\begin{enumerate}
    \item 50 games
    \item 60 plays
    \item 52142 frames (frame means image)
\end{enumerate}
        \begin{figure}[h]
        \centering
        \includegraphics[width=100mm]{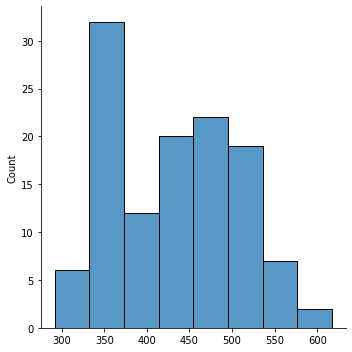}
        \caption{Frame per game}
        \label{fig:frame_per_game}
        \end{figure}
        
\item How are the mapping done \ref{fig:game_x_y}
        \begin{figure}[h]
        \centering
        \includegraphics[width=100mm]{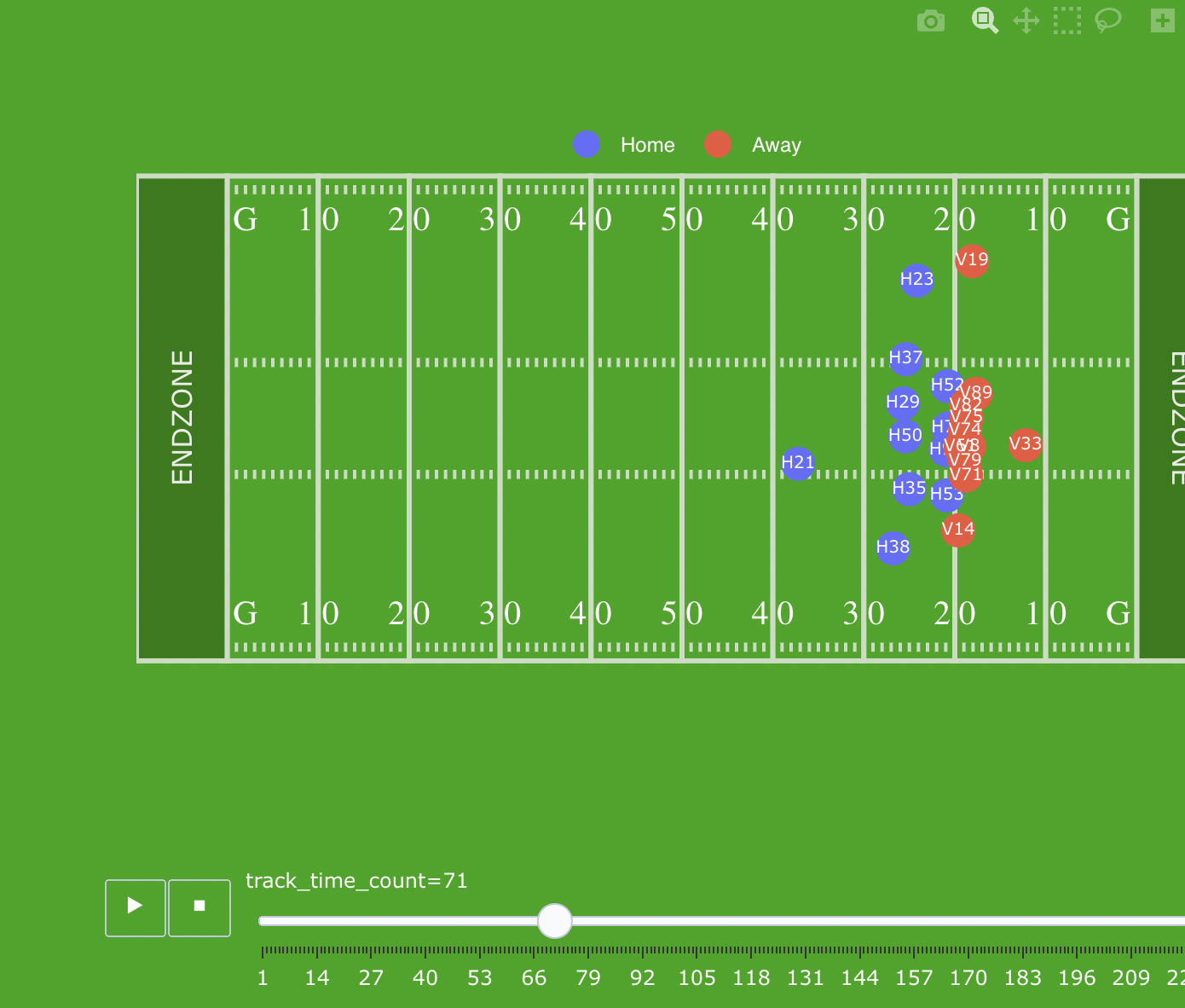}
        \caption{Game Mapping in X Y coordinate}
        \label{fig:game_x_y}
        \end{figure}
        
\item Helmet information:  This feature doesn't seem useful to our competition \ref{fig:helmet_info}
        \begin{figure}[h]
        \centering
        \includegraphics[width=100mm]{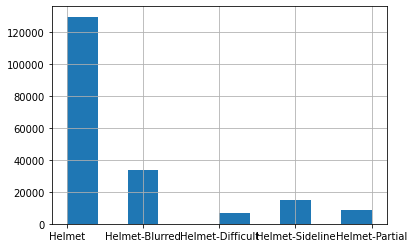}
        \caption{Plot for different kind of helmets}
        \label{fig:helmet_info}
        \end{figure}
        
\item How are images seen with very low confidence in bounding Box \ref{fig:image_high_conf}
        \begin{figure}[h]
        \centering
        \includegraphics[width=100mm]{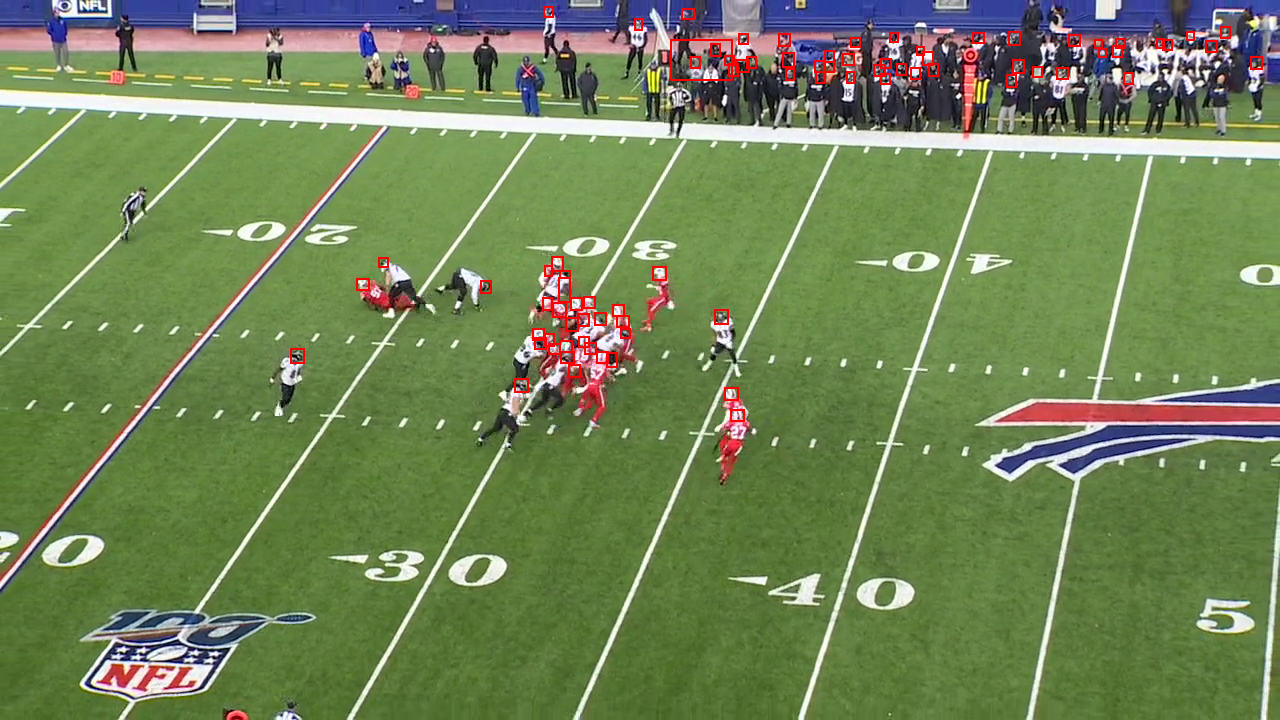}
        \caption{very Confidence Bounding Box}
        \label{fig:image_high_conf}
        \end{figure}\\
\item How are images seen with very a little high confidence in bounding Box \ref{fig:low_conf_image}
        \begin{figure}[h]
        \centering
        \includegraphics[width=100mm]{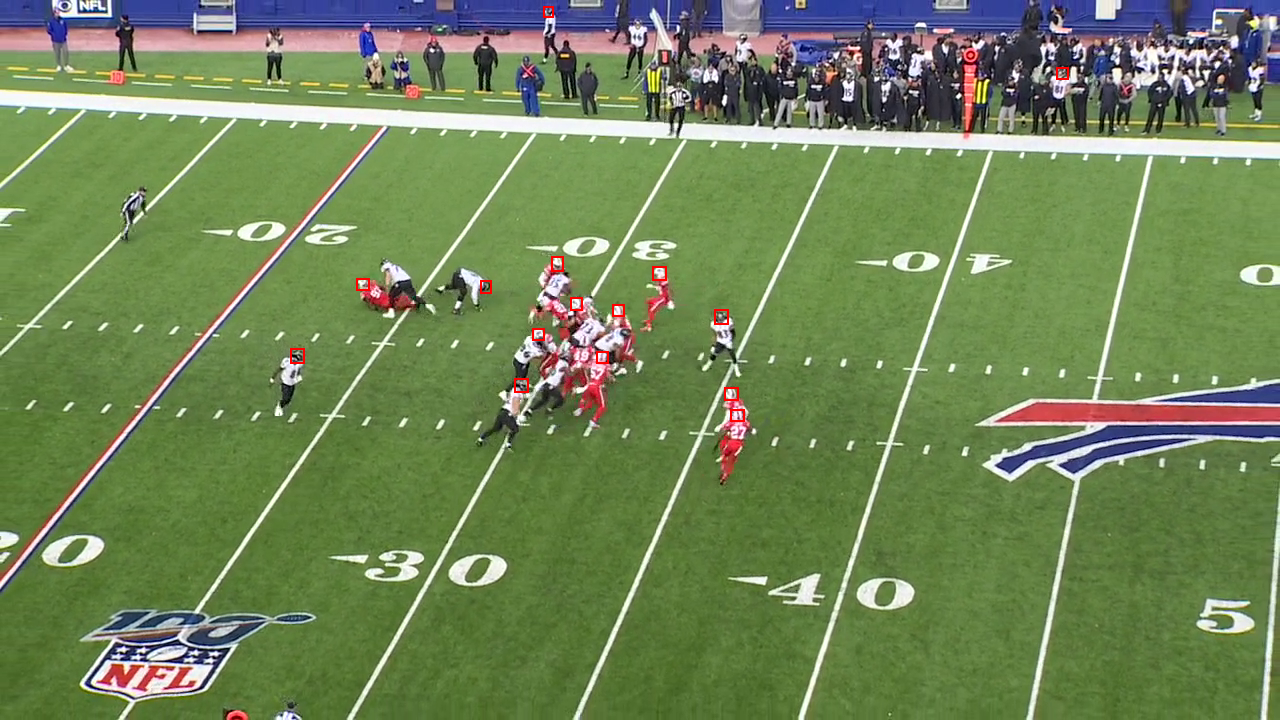}
        \caption{High Confidence Bounding Box}
        \label{fig:low_conf_image}
        \end{figure}\\
\item How are images with tracking information seen?  \ref{fig:tracking_info}
        \begin{figure}[h]
        \centering
        \includegraphics[width=100mm]{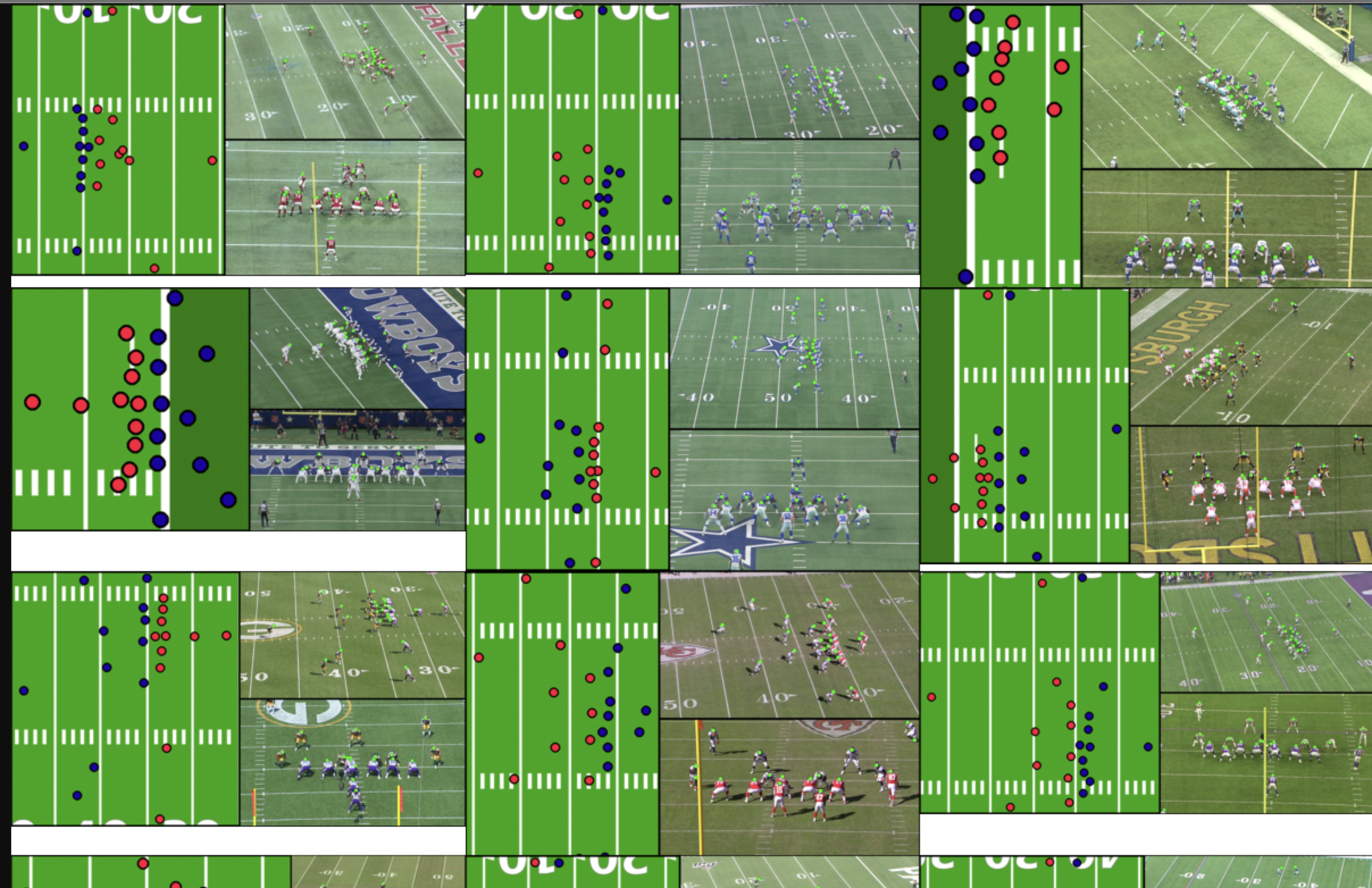}
        \caption{Mapped image with tracking information}
        \label{fig:tracking_info}
        \end{figure}\\
\end{enumerate}

\subsection{Modeling for Detectron2}
We tried detectron for our modeling as well, but unfortunately we couldnt implement it due to constraints in time and resources, we are sure it would have made a lot of improvements. We will list the procedures here for further references: 
\begin{figure}[h]
    \centering
    \includegraphics[width=100mm]{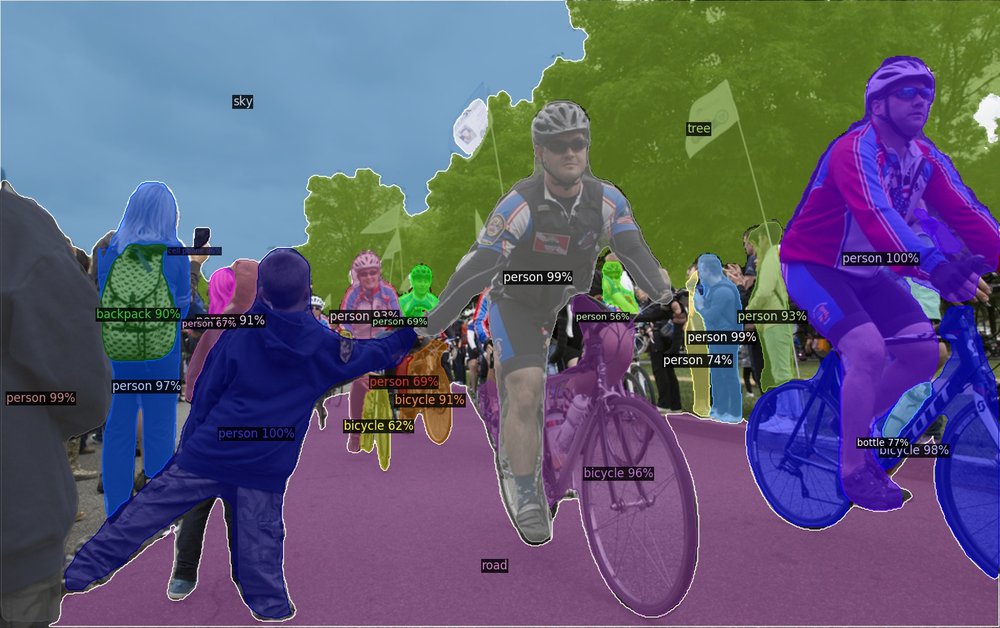}
    \caption{Detectron}
    \label{fig:detectron2}
    
\end{figure}

we took all the references from \cite{wu2019detectron2}. 
Detectron2 is Facebook AI Research's next generation library that provides state-of-the-art detection and segmentation algorithms. It is the successor of Detectron and maskrcnn-benchmark. It supports a number of computer vision research projects and production applications in Facebook.

Official tutorial in detectron2 use their API and need to write custom training script, but we prefer to use default CLI interface. To do that, we need to dig into the code where datasets are set up and register our custom dataset information. Our code show how to modify $detectron2/data/datasets/builtin.py$ to register $nfl2021\_train$ and $nfl2021\_valid$ dataset.
\subsubsection{Config Setup}
We gave an example to use FasterRCNN, but you can use any of configs in detectron2 repository. \cite{MODEL_ZOO_DETECTRON2}

Below looks much simpler than mmdetection configs, but it actually has a lot of hidden parameters in default config. (Personally we felt it's more explicit to expose all config in yaml file.) below are the summary of what you need to modify.

\begin{enumerate}
    \item Dataset: You need to specify dataset name both for train and valid. The details are already registered, so you don't need to set here.
    \item Model: You need to change num classes($MODEL.ROI\_HEADS.NUM\_CLASSES$) to 1.(We only need to detect only Helmet class.)
    \item Others: You need to set pretrained model path in MODEL.WEIGHTS because what we want to do here is fine-tune, not training from scratch.
Also it's better to decrease LR for fine tuning.
You can also change max\_epochs as you want, here I set to 1 for just demonstration.
\end{enumerate}

\subsubsection{Dataset}
We had to construct a custom dataset in the form of COCO format which is needed by detectron2 to run, since the default format for our original dataset is csv, we had to convert this dataset into a json format as seen in Figure \ref{fig:coco}

\begin{figure}[h]
    \centering
    \includegraphics[width=150mm]{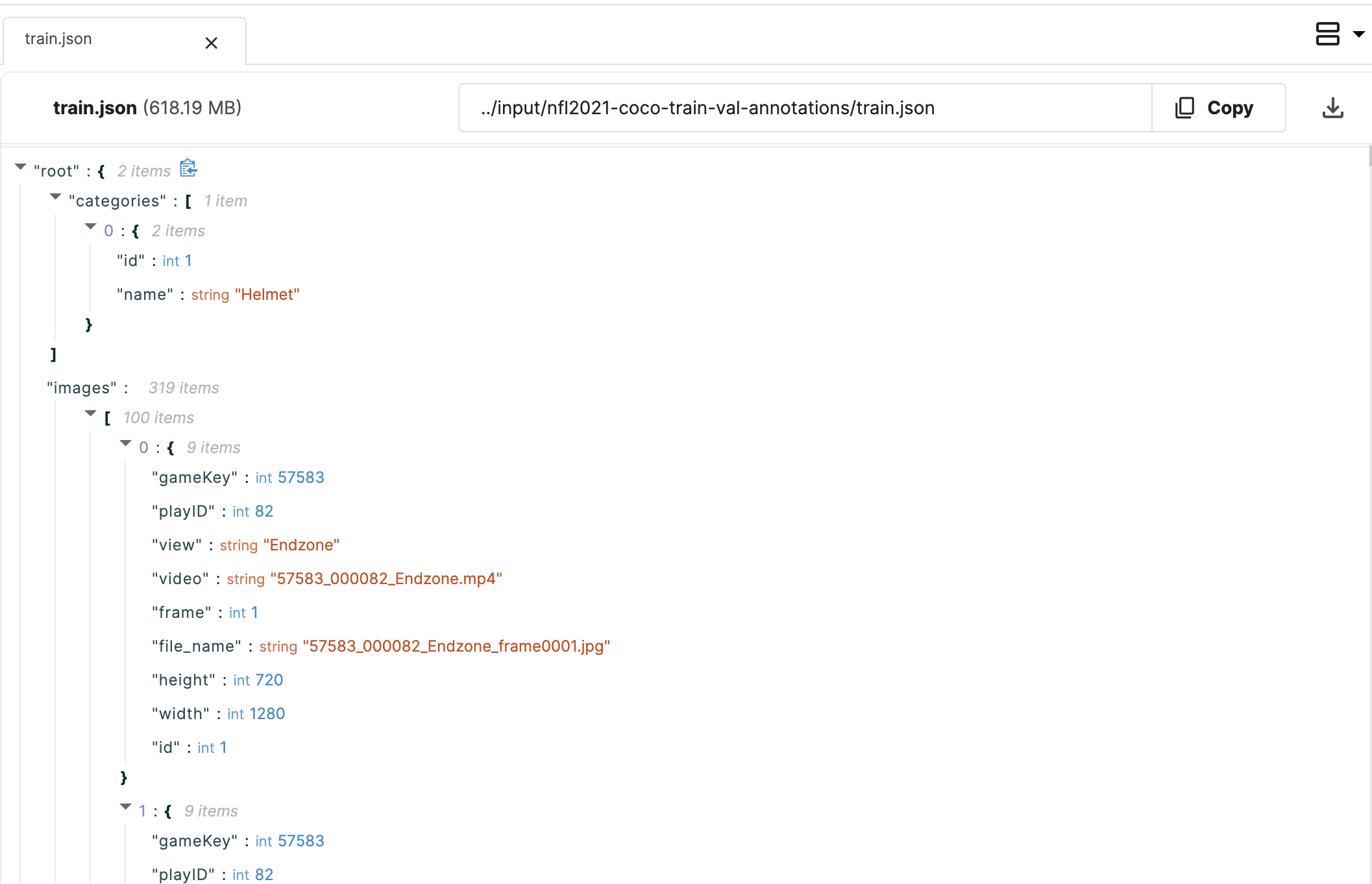}
    \caption{Detectron COCO dataset format}
    \label{fig:coco}
\end{figure}
\subsubsection{Results}
We were able to obtain pretty good results on bounding boxes for our images, while we could not do it for videos we are sure this can be a very powerful way for detecting in future competitions. We explored both segmentation and detection of bounding boxes using DETR2, and found good results in this. However, the training time take for 9000 images was more than expected and many times Kaggle notebooks used to crash because of high utilisation of GPUs.
\begin{figure}[h]
    \centering
    \includegraphics[width=150mm]{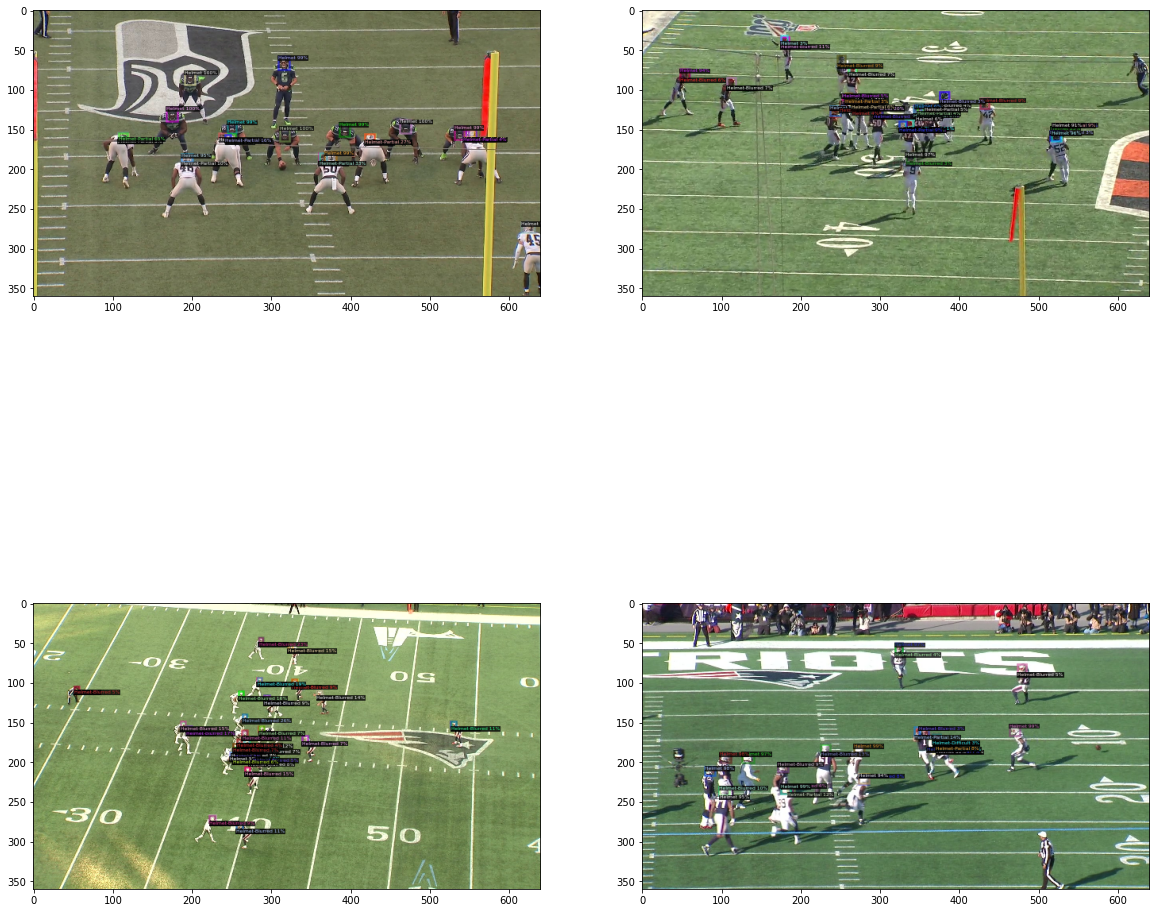}
    \caption{Output Images}
    \label{fig:coco}
\end{figure}

\begin{figure}[h]
    \centering
    \includegraphics[width=150mm]{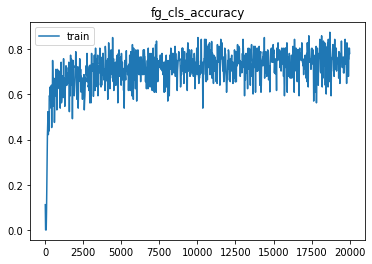}
    \caption{Class Accuracy}
    \label{fig:coco}
\end{figure}
\begin{figure}[h]
    \centering
    \includegraphics[width=150mm]{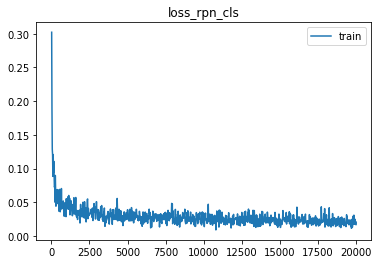}
    \caption{Loss curve}
    \label{fig:coco}
\end{figure}

\end{document}